\documentclass[runningheads]{llncs}
\usepackage[T1]{fontenc}
\usepackage{graphicx}
\usepackage{algorithm}
\usepackage{algpseudocode}
\usepackage{multicol}
\usepackage{multirow}

\begin{document}

\title{Syntactic Complexity Identification, Measurement, and Reduction Through Controlled Syntactic Simplification}
\author{Muhammad Salman\inst{1}\orcidID{0000-0003-4417-7994} \and
Armin Haller\inst{1}\orcidID{0000-0003-3425-0780} \and
Sergio J. Rodríguez Méndez\inst{1}\orcidID{0000-0001-7203-8399}}
\authorrunning{M. Salman et al.}
\institute{School of Computing - CECC, The Australian National University, 2601 Acton, Australia\\
\url{https://www.anu.edu.au/about}}
\maketitle              % typeset the header of the contribution
*Corresponding Author: muhammad.salman@anu.edu.au (Salman M.)

\begin{abstract}
Text simplification is one of the domains in Natural Language Processing (NLP) that offers an opportunity to understand the text in a simplified manner for exploration. However, it is always hard to understand and retrieve knowledge from unstructured text, which is usually in the form of compound and complex sentences. There are state-of-the-art neural network-based methods to simplify the sentences for improved readability while replacing words with plain English substitutes and summarising the sentences and paragraphs. In the Knowledge Graph (KG) creation process from unstructured text, summarising long sentences and substituting words is undesirable since this may lead to information loss. However, KG creation from text requires the extraction of all possible facts (triples) with the same mentions as in the text. In this work, we propose a controlled simplification based on the factual information in a sentence, i.e., triple. We present a classical syntactic dependency-based approach to split and rephrase a compound and complex sentence into a set of simplified sentences. This simplification process will retain the original wording with a simple structure of possible domain facts in each sentence, i.e., triples. The paper also introduces an algorithm to identify and measure a sentence's syntactic complexity (SC), followed by reduction through a controlled syntactic simplification process. Last, an experiment for a dataset re-annotation is also conducted through GPT3; we aim to publish this refined corpus as a resource. This work is accepted and presented in IWLKG\footnote{International Workshop On Learning With Knowledge Graphs @ Web Search and Data Mining (WSDM)} workshop at WSDM\-2023 Conference. The code and data is available at \url{https://github.com/sallmanm/SynSim}.


\end{abstract}

\section{Introduction}
In linguistics, a natural language is used to be evolved naturally in humans with the use and repetition and without conscious planning or premeditation \cite{lyons_1991}. Natural languages used to have various forms, such as speech or signing. Textual data present these days is mainly in natural language text, with compound and complex structures and unstructured forms. In automatic text processing and information extraction systems, unstructured text restricts achieving desired performance. It is because most IE systems need accurate and complete knowledge identification and extraction. However, these systems perform efficiently on plain/simple language text \cite{niklaus2018surveyOpenIE}.

The complexity of an English language sentence depends mainly on the clausal structure and the length of the sentence \cite{chen2011computingSCfeatures}. Based on this information, a sentence can be categorized in one of the following:
\begin{enumerate}
    \item Simple Sentence: It includes an independent clause. There is no long sentence structure.
    \item Compound Sentence: A sentence with two or more coordinated clauses form a compound structure that can be long, difficult to read and track.
    \item Complex Sentence: Sentences that contain at least one coordinated and one subordinated clause are in the category of complex sentences.
    \item Compound-Complex Sentences: These are the most complex sentences regarding reading and tracking. This type of sentence comprises more than one subordinate and coordinate clause.
\end{enumerate}

Table-\ref{tab:Sent-Structure} presents the structure of sentences following a simple statistical analysis. Based on these signs \cite{evans2019identifyingSignsSC}, we can identify and classify the sentence in one of the categories, followed by the complexity reduction approach. In the following section, we propose an algorithm to identify the nature of a sentence, whether it is simple or not.

\begin{table}[ht]
\centering
\caption{Sentences classification based on their nature}
\label{tab:Sent-Structure}
\begin{tabular}{c|cccc|}
\cline{2-5}
\multirow{2}{*}{} & \multicolumn{2}{c|}{\textit{\textbf{Clause}}} & \multicolumn{2}{c|}{\textit{\textbf{Tokens}}} \\ \cline{2-5} 
 & \multicolumn{1}{c|}{Coordinate} & \multicolumn{1}{c|}{Subordinate} & \multicolumn{1}{c|}{Length} & \multicolumn{1}{c|}{Verbs} \\ \hline
\multicolumn{1}{|c|}{Simple} & 0 & 0 & \textless{}=8 & 1 \\ \cline{1-1}
\multicolumn{1}{|c|}{Compound} & \textgreater{}=2 & 0 & \textgreater{}10 & \textgreater{}=2 \\ \cline{1-1}
\multicolumn{1}{|c|}{Complex} & 1 & \textgreater{}=1 & \textgreater{}10 & \textgreater{}=2 \\ \cline{1-1}
\multicolumn{1}{|c|}{Compound-Complex} & \textgreater{}=2 & \textgreater{}=1 & \textgreater{}10 & \textgreater{}=2 \\ \cline{1-5}
\end{tabular}
\end{table}

\subsection{Motivation and Problem Statement}
Information Extraction (IE) is the task of retrieving information from data (this work deals with unstructured text), and it makes the machine capable of understanding and comprehending textual data \cite{vo2017openIE}. However, there are still specific challenges for IE systems when dealing with unstructured text \cite{niklaus2018surveyOpenIE}. The syntactic structure of a sentence is vital to achieving higher performance in IE tasks, i.e., the more simplified sentence, the more efficient the IE system. In the recent approaches, it has been identified intuitively that short sentences are easy to process for many information retrieval tasks \cite{bahdanau2014neuralMT,song2018leveragingQG,ondov2022surveyBIOss}. Similarly, the performance of triple (subject, predicate, object) extraction systems is also better when the input text is cleaned, simple, and short \cite{van2021mayIhelp}. In the task of triple extraction, Syntactic Complexity (SC) plays a vital role in the performance of an IE system. It can be reduced while transforming the sentence into simple sentences. In section \ref{ST}, we present the \emph{Syntactic Simplification (SynSim)} process, which retains the actual mentions of an input sentence, and each simplified sentence contains at least one triple (subject, verb, object).
Moreover, we have proposed an algorithm that provides a score based on the identified syntactic signs of the sentence. We can classify the simple and complex sentences with a threshold, followed by the SynSim process for complex sentences. Table-\ref{tab:Sent-Triplification} shows an example of our work where a compound and complex sentence is transformed into a set of controlled simplified sentences, i.e., each simplified sentence retains at-least one triple after clauses' splitting and rephrasing. 

\begin{table*}[ht]
\centering
\label{tab:Sent-Triplification}
\caption{Example of Complex to Simplified Sentences}
\begin{tabular}{|c|c|c|c|c|}
\cline{1-1}
{\textbf{COMPOUND \& COMPLEX SENTENCE}}\\ \cline{1-1}
FIFA World Cup 2022 is taking place in Qatar, the first to be held in the Arab world,\\ and the second held in entire Asia.  \\ \cline{1-1}
{\textbf{Syntactically Simplified SENTENCES}}  \\ \cline{1-1}
FIFA World Cup 2022 is taking place in Qatar.  \\\cline{1-1}
FIFA World Cup 2022 is the first to be held in the Arab world .   \\\cline{1-1}
FIFA World Cup 2022 is the second held in entire Asia.   \\\cline{1-1}
\end{tabular}
\end{table*}

\section{Literature Review}
Sentence simplification has been addressed in linguistics since the 1990s \cite{agarwal1992simple}. This section briefly reviews the sentence simplification and transformation methods, specifically syntactically exploited approaches. The lexical structure-based simplification approaches are unrelated to our work, so we do not cover the review of those methods. Syntactic structure-based text simplification started with rule-based approaches \cite{al2021automatedSS} have facilitated many information extraction systems, i.e., information retrieval for biomedical textual data \cite{ondov2022surveyBIOss}, dependency parsing \cite{agarwal1992simple}, and semantic labelling \cite{vickrey2008labelling}. Exploiting pre-processing, \cite{siddharthan2006syntactic} proposed a handcrafted patters based approach which evolves tokenization, Part of Speech (PoS) tagging, and chunking to analyze the input text. To improve the analysis of the input text, \cite{evans2011comparing} integrated a machine learning (ML) based approach to classify the tagging and chunking. 

Working with the sentence's syntactic structure, many techniques are evolved to simplify the sentence by re-ordering and re-writing the complex/compound sentences \cite{max2000syntactic}. Multilingual Syntactic Simplification Tool (MUSST) \cite{scarton2017musst} assists in interacting with various organizations and people with different natural languages. This tool has evolved and improved the machine translation approaches. MUSST claims 70\% accuracy in simplifying English language sentences that belong to the PA (Public Administration) textual data \cite{scarton2017musst}.

Statistical ML-based approaches evolved after the availability of simplified data i.e., Plain/Simple Wikipedia \cite{coster2011simple} for English language articles. ML-based approaches rely on original and transformed (Simple) text, but it is not currently available for all the contexts of texts. Due to this limitation, ML-based approaches are practical for only available context-oriented data \cite{alva2020datadrivenSS}. For the available textual articles of Wikipedia and Simple Wikipedia, \cite{siddharthan2006syntactic} derived the probability table of n-grams transformation for these articles and provided a machine translation model to transform text in simplified form. \cite{wubben2012sentence} also applied this approach, and after capturing the intuitions (phrase table), the model transformed the sentence by performing phrase deletion/re-ordering and lexical substitution of words. DRESS (Deep Re-Enforcement Sentence Simplification )\cite{zhang2017DRESS} applied neural machine translation methods to train on original and manually simplified textual data. DRESS affected the structure of the simplified sentence to meet the constraints of simplicity i.e., re-ordering, deletion, and substitution. In \cite{evans2019identifyingSignsSC}, the author proposed an approach that does not require any corpus or phrase table to simplify the sentence but applies an iterative rule-based approach to identify the complexity signs and handles with the reconstruction of a sentence.

In this work, our approach does not rely on data corpus or handcrafted rules but a measured score of complexity \ref{SynComplex}, followed by a simplification process with splitting and rephrasing \ref{ST}. Furthermore, our approach is iterative and does not include any substitution and deletion but contains the original wording and textual mentions.

\section{Proposed Methodology}
This section discusses our approach to identifying a sentence’s complexity and transforming it into a set of simplified sentences. Their evaluation is also included in the following respective sections.

\subsection{Syntactic Complexity Identification and Classification}\label{SynComplex}
The sentence structure that reduces readability and increases textual ambiguity for explainable Natural Language Understanding (NLU) is known as complexity, which is classified into lexical and syntactic aspects. \emph{Lexical complexity} refers to the form of complexity in which complex wording and ambiguous/unknown grammatical structure are present, whereas \emph{Syntactic complexity} refers to the clausal structure and length of the sentence \cite{al2021automatedSS,jagaiah2020syntacticWriting}. Clauses in a sentence are based on the conjunctions which link multiple clauses to form one single multi-clause sentence \cite{evans2019identifyingSignsSC}.

Sentences with conjunctions can be labelled as compound and complex with a scoring approach, which will consider three main components of a sentence i.e., the number of conjunctions, verbs, and tokens. In the past, approaches to identify and mark the signs of complexity have been proposed, and the main contribution in this area of research is by Evans and Orasan et al. \cite{evans2011comparing,evans2014evaluation,evans2019identifyingSignsSC}. Some other approaches deal with syntactic complexity in the domain of linguistics writing and understanding \cite{jagaiah2020syntacticWriting}.
A complex sentence can be rephrased into a set of simple sentences so that information extraction systems can efficiently process it to identify the facts accurately.
Conjunction could be either coordinated or subordinated. In a compound sentence, coordinated conjunctions are present, which connects two or more independent clauses with “for, and, nor, but, or, yet, so, and semicolon”. A complex sentence structure slightly differs from compound sentences as it constitutes a dependent clause and one relative independent clause.

\textbf{Proposed Algorithm:}
We have proposed Algorithm-\ref{alg:SC} to measure the \emph{Syntactic Complexity (SC)} of a sentence. With an SC threshold, we can identify a sentence's state and complexity level.
 
\begin{algorithm*}
\caption{Syntactic Complexity Measurement Algorithm}\label{alg:SC}
\begin{algorithmic}[1]
\State \textbf{Input:} $Sentence (SENT), Syntactic Complexity (SC)=0$
\State Token Weight: $ T_W = 0.07 $ 
\State Verb Weight: $ V_W = 0.3 $ 
\State Conjunction Weight: $ C_W = 0.4 $ 
\State \textbf{Pre-process} $\gets Tokenizer|POS\_Tagger$ 
\State $SENT \gets Preprocess(SENT)$
\State $TKN \gets LEN(SENT)$            \Comment{Total Number of Tokens} 
\State  $CNJ , VRB \gets 0$         \Comment{Initializing with 0} 
\For{\texttt{Token in SENT}}
    \If{$Token.Tag == VERB$}
        \State $VRB \gets +1$           \Comment{Adding 1 in VERB Count} 
    \ElsIf{$Token.Tag == Conjunction$}
        \State $CNJ \gets +1$           \Comment{Adding 1 in Conjunction Count} 
        \Else
            \State $CONTINUE$
    \EndIf
\EndFor
\State $SC \gets TKN*W_T+VRB*W_V+CNJ*W_C$
\State \textbf{Return SC}

\end{algorithmic}
\end{algorithm*}

\subsection{Controlled Sentence Simplification} \label{ST}
\emph{Syntactic Simplification (SynSim)} is the process of splitting and rephrasing a compound sentence based on independent clauses it contains. SynSim generates a set of simplified sentences where each simplified sentence has one independent clause. Our objective is to transform a complex sentence into a set of sentences with minimal syntactic complexity; and with at least one fact mentioned (subject, verb, object) in sub-sentences. Automatic SynSim will help to improve the readability of raw and unstructured text for machines to extract meaningful information. It will also help Knowledge Graph (KG) creation methods perform better in extracting RDF triples. It is because SynSim does not include the approaches of updating words with synonyms and contains the exact predicate's mention, which will be helpful in the identification of extracted triples in a target KG. \\

\begin{algorithm}[ht]
\caption{Sentence Splitting and Rewriting Algorithm}\label{alg:SynSim}
\begin{algorithmic}[1]
\State \textbf{Input:} $Sentence (SENT), Syntactic Complexity (SC)$
\State \textbf{Output:} $Simple\_Sentences(SS)\gets\emptyset$
\While{$SC(SENT)>Threshold$}
\State $s_1,s_2,s_3,...,s_n \gets SynSim(SENT)$
\State $\textbf{Push}(s_1,s_2,s_3,...,s_n,SENT)$
\EndWhile
\For{s in SENT}
    \If{$SC(s)>Threshold$}
    \State $s_1,s_2,s_3,...,s_n \gets SynSim(s)$
    \State $\textbf{Push}(s_1,s_2,s_3,...,s_n,SS)$
    \Else
    \State $\textbf{Push}(s,SS)$
    \EndIf
\EndFor
\State \textbf{Return} Simple\_Sentences(SS)

\end{algorithmic}
\end{algorithm}

\begin{figure*}[ht]\label{SynSimExample}
\centering
\caption{Example of Clause Identification and Rephrasing\\}
\includegraphics[width=0.7\linewidth]{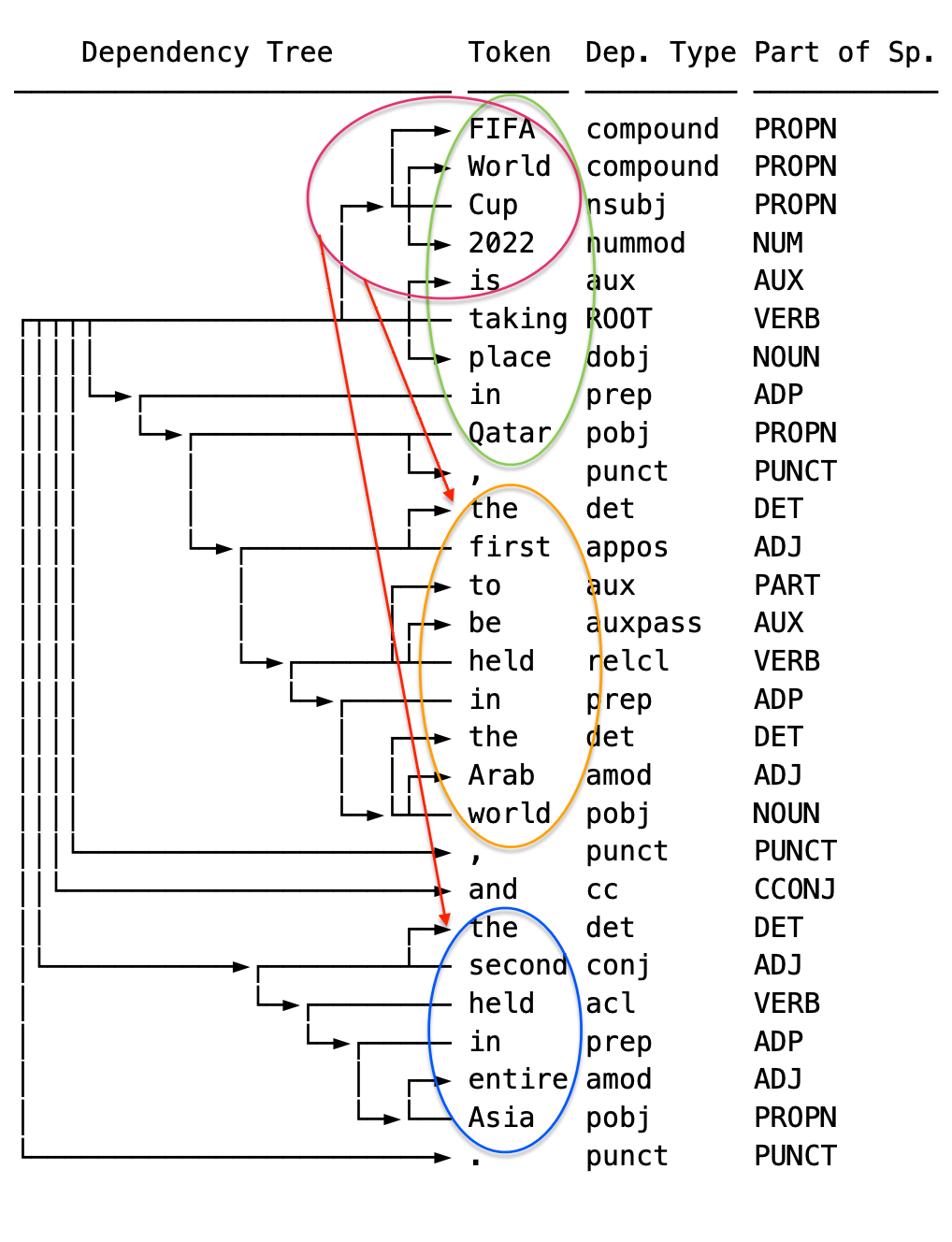}
\label{STExample}
\end{figure*}

SynSim involves the process of conjunction identification and \emph{Entity Co-reference Resolution (ECR)}\footnote{Task of identification and replacement of textual mentions which refer to the same entity in text}. Based on the conjunctions, clauses are identified, followed by a splitting function to get a set of independent clauses of a sentence. In the last step, ECR rephrases the clauses with respective entities' co-reference. The example of the SynSim process is shown in Figure-\ref{SynSimExample} where three (3) independent clauses are identified by the SynSim and circled (Green, Yellow, Blue). The red circled phrase is the identified parent entity and refers to all the sentence clauses; it will be linked to the sub-clauses in the process of ECR.

In terms of controlled SynSim, we have integrated the process of SynSim with SC score to avoid high computational costs for all the sentences. This approach uses a threshold to classify the simplified sentences against compound and complex sentences. A complex sentence will remain in the loop of 'splitting and rephrasing' until all the sub-sentences are labelled as simplified i.e., $SC<Threshold$. The pseudo-code of the proposed approach is shown in Algorithm-\ref{alg:SynSim}.

\section{Performance and Relevance}
\subsection{Performance of SC Identification} 
 We have evaluated our proposed Algorithm-\ref{alg:SC} on a manually annotated corpus\cite{zhang2020SmallButMighty} \emph{'Spilt \& Rephrase'\footnote{https://developer.ibm.com/data/split-and-rephrase/}}, launched by IBM. This dataset contains 720 complex sentences, which are transformed into a set of simple sentences by annotators. We adjusted the hyper-parameters of our model with an SC threshold of 1 i.e. if $SC > 1$, then the sentence will be labelled as complex and simple otherwise.

In dependency tree-based SC measurement, we considered only five dependency tags which include 'ROOT', 'ATTR', 'CC', 'CONJ', and 'PUNCT(,)'. We grouped them into a prescribed category i.e., 'VRB: VERB' and 'CNJ: Conjunction'. The third variable in our algorithm is 'TKN: Tokens/Terms', which is a count of the primary tokens/words of a sentence while excluding the stop-words. The performance of dependency-based SC methods is nearly ideal, as shown in Table 3.

We have also applied a Part-of-speech (POS) based SC measurement algorithm on the same dataset. We used just three tags in this experiment: ' VERB', 'CCONJ', and 'PUNCT'. POS-based SC experiment was also successful as it has also achieved outstanding accuracy. \\

\begin{table*}[ht]
\label{performSC}
\centering
\caption{Performance of Proposed Algorithm}
\begin{tabular}{|c|cc|cc|}
\hline
\multirow{2}{*}{} & \multicolumn{2}{c|}{{Dep.Tree-Based SC}}  & \multicolumn{2}{c|}{\textit{\textbf{POS-Based SC}}}\\\cline{2-5}
                  & \multicolumn{1}{c|}{Complex}           & {Simple}          & \multicolumn{1}{c|}{Complex}        & {Simple}        \\ \hline
Complex           & \multicolumn{1}{c|}{690}               & {30}               & \multicolumn{1}{c|}{689}            & {31}             \\ \hline
Simple            & \multicolumn{1}{c|}{605}                 & {2960}             & \multicolumn{1}{c|}{548}             & {3017}           \\ \hline
\end{tabular}
\end{table*}

\subsubsection{Validation and Refinement of Dataset}
After analysing the misclassified simple sentences as complex sentences, we have investigated that the simple version of the sentence is extended and compound. We found nearly 650 sentences out of 3565 simplified sentences as long and compound. This dataset was meant to provide simple sentences but eventually ended up with more sentences while containing nearly the same number of compound sentences as the original dataset started with i.e., 720 total compound and complex sentences. After this investigation, we re-annotated the dataset with simple sentences through very recently launched \emph{GPT3} by OpenAI\footnote{https://beta.openai.com/} and discussed it in the following section.

\subsection{Simplification through \emph{GPT3} and Human Evaluation}
In the annotation process, we integrated \emph{ChatGPT} for the process of \emph{'Split and rephrase'}. We applied the most powerful model of GPT3 i.e., Text-Davinci-003, to annotate the complex sentence with simple sentences. After getting the annotation of GPT3, we manually examined and validated the annotated simple sentences. We did not remove or modify any sentence to keep the original output of GPT3 for SC measurement. Against 720 complex sentences, GPT3 returned us with 2277 syntactically modified simple sentences. SC evaluation for the re-annotated dataset is shown in Table 4. 
\begin{table*}[ht]
\label{SC1}
\centering
\caption{Performance of Proposed Algorithm On Re-Annotated dataset with GPT3}
\begin{tabular}{|c|cc|cc|}
\hline
\multirow{2}{*}{} & \multicolumn{2}{c|}{{Dep. Tree-Based SC}} & \multicolumn{2}{c|}{\textit{\textbf{POS-Based SC}}} \\ \cline{2-5} 
                  & \multicolumn{1}{c|}{Complex}           & Simple          & \multicolumn{1}{c|}{Complex}        & Simple        \\ \hline
Complex           & \multicolumn{1}{c|}{685}               & 35               & \multicolumn{1}{c|}{691}            & 29             \\ \hline
Simple            & \multicolumn{1}{c|}{262}                 & 2015             & \multicolumn{1}{c|}{166}             & 2111           \\ \hline
\end{tabular}
\end{table*}

Furthermore, after assessing the performance in Table 5, we acknowledge this dataset still has a margin for simplification, and we will look into that for our resource paper. We plan to publish a new refined dataset with verified simple sentences.

\begin{table*}[ht]
\caption{Performance Statistics of SC Measurement}
\centering
\label{PerformanceSC1}
\begin{tabular}{|cl|l|}
\hline
\multicolumn{2}{|c|}{\textbf{Description}} & \textbf{Value} \\ \hline
\multicolumn{2}{|c|}{Complex Sentences} & 720 \\ \hline
\multicolumn{1}{|c|}{\multirow{2}{*}{Simplified Sentences}} & IBM Corpus & 3565 \\ \cline{2-3} 
\multicolumn{1}{|c|}{} & GPT3 Re-annotated & 2277 \\ \hline
\multicolumn{1}{|c|}{\multirow{2}{*}{Avg. Simplified Per Sentence}} & IBM Corpus & 4.95 \\ \cline{2-3} 
\multicolumn{1}{|c|}{} & GPT3 Re-annotated & 3.16 \\ \hline
\multicolumn{1}{|c|}{\multirow{2}{*}{Performance}} & Accuracy IBM Corpus & \textbf{90.15\%} \\ \cline{2-3} 
\multicolumn{1}{|c|}{} & Accuracy GPT3 Corpus & \textbf{94.34\%} \\ \hline
\end{tabular}
\end{table*}

\subsection{Relevance of Syntactic Simplification (SynSim)}
We have evaluated the SynSim approach on the \emph{'Split \& Rephrase'\footnote{https://developer.ibm.com/data/split-and-rephrase/}} corpus, which we used for SC evaluation as well. We compared the relevance of SynSim's output with simple sentences of the annotated corpus. The similarity is calculated with two evaluation metrics i.e., \textit{Cosine Similarity} and \textit{Jaccard Similarity} as stated in Equation-\ref{cosine} and Equation-\ref{jaccard} respectively. Two sets of sentences are formed for each corpus output and SynSim against a complex sentence. For example, suppose $O_{corpus}$ and $O_{SynSim}$ be two sets of sentences for annotated corpus and SynSim output, and each set is formed as [ $O_{corpus}\ \forall \ sent \ in\ Simplified\_Sentences$].

\begin{equation}\label{cosine}
  Cosine\_Similarity(O_{corpus},O_{SynSim}) = \frac{O_{corpus} \cdot O_{SynSim}}{|O_{corpus}||O_{SynSim}|}  
\end{equation}

\begin{equation}\label{jaccard}
  Jaccard\_Similarity(O_{corpus},Simp) = \frac{|O_{corpus} \cap O_{SynSim}|}{|O_{corpus} \cup O_{SynSim}|}  
\end{equation}
\\
%%%
In our evaluation results, our method \emph{SynSim} produced 2364 simplified sentences out of 720 compound and complex sentences. Comparing it with the annotated corpus, SynSim produced almost 33\% less simplified sentences than the annotated corpus contains. The more simplified sentences, the more chances for the information extraction system to perform better. Our system produced more than the actual annotations in this metric, which is an advantage.  
%%%
In the next step, we investigated the relevance of our simplified sentences with the actual annotated corpus. To identify relevance, we applied \emph{Cosine Similarity} and \emph{Jaccard Similarity} as stated in Equation-\ref{cosine} and Equation-\ref{jaccard}, respectively. The output relevance scores are shown in Table \ref{PerformanceSynSim}.

\begin{table*}[ht]
\caption{Performance Statistics }
\centering
\label{PerformanceSynSim}
\begin{tabular}{|cl|l|}
\hline
\multicolumn{2}{|c|}{\textbf{Description}} & \textbf{Value} \\ \hline
\multicolumn{2}{|c|}{Complex Sentences} & 720 \\ \hline
\multicolumn{1}{|c|}{\multirow{2}{*}{Simplified Sentences}} & Corpus & 3565 \\ \cline{2-3} 
\multicolumn{1}{|c|}{} & SynSim & 2364 \\ \hline
\multicolumn{1}{|c|}{\multirow{2}{*}{Avg. Simplified Per Sentence}} & Corpus & 3.015 \\ \cline{2-3} 
\multicolumn{1}{|c|}{} & SynSim & 3.283 \\ \hline
\multicolumn{1}{|c|}{\multirow{2}{*}{RELEVANCE}} & Cosine & \textbf{94.53\%} \\ \cline{2-3} 
\multicolumn{1}{|c|}{} & Jaccard & \textbf{90.42\%} \\ \hline

\end{tabular}
\end{table*}

\section{Conclusion}
The proposed system involves two models that identify and resolve the textual complexity issue. In measuring sentence complexity, we used syntactic signs of linguistics and calculated the score against each present sign in the sentence. The weight of each sign is carefully monitored and tailored so that it can successfully classify complex and simple sentences with a threshold. We evaluated our complexity measurement system with a threshold and tested it on a benchmark corpus. After investigating the 'Split \& Rephrase' corpus, we identified that the simplified sentences are complex, but the dataset still has as many complex sentences as the original input data had. Therefore, we refined this corpus by evolving the recently launched GPT3 model. We evaluated the performance of our SC system on both versions of the corpus and plan to publish the resource paper with refined and verified annotations.\\
Following the complexity measurement, our second model reduced the complexity by splitting and rephrasing the identified complex sentence. In the complexity reduction process, we applied controlled syntactic simplification in which we retained the actual mentions' of the text. Finally, we evaluated \emph{Syntactic Simplification (SynSim)} model with the 'Cosine' and 'Jaccard' similarity metrics. SynSim achieved 94.5\% and 90.4\% similarity scores while evaluating the relevance of output with COSINE and JACCARD similarity, respectively. We have also applied the Triple (Subject, Verb, Object) method on simplified sentences and plan to publish a complete pipeline from unstructured text to triple extraction for Knowledge Graphs.
%%%

\textbf{Key Features:}
\begin{enumerate}
    \item Splitting and rephrasing will not lose any of existing triples.
	\item It will not modify the actual mentions of triples.
	\item It produces sentences with minimal syntactic complexity.
	\item Predicates are not replaced with simplified synonyms to retain the actual predicate’s URI while entity identification task.
\end{enumerate}

\textbf{Purpose and Benefit:}
\begin{enumerate}
    \item SC measurement will help the approach to classify before SynSim as it could reduce the computational cost and resources' utilization.
    \item The purpose of controlled SynSim is beyond the task of summarizing, and readability, however, this method will be helpful to improve the machine readability, knowledge extraction, and automatic RDF triple extraction from unstructured text.
	\item It will improve the performance of Knowledge Graph (KG) creation methods for unstructured text.
	\item It will also help to identify the exact triples’ mentions in a target KG and eventually will be helpful for knowledge identification, completion, and enrichment.
\end{enumerate}
%%%

\bibliography{complexity.bib}
\bibliographystyle{unsrt}

\end{document}